\pdfoutput=1

\documentclass[11pt]{article}

\usepackage[]{EMNLP2022}

\usepackage{graphicx}
\usepackage{times}
\usepackage{latexsym}
\usepackage{enumitem}
\usepackage{amsmath}
\usepackage{float}

\usepackage[T1]{fontenc}

\usepackage[utf8]{inputenc}

\usepackage{microtype}

\usepackage{inconsolata}

\title{Using Deep Mixture-of-Experts to Detect Word Meaning Shift for TempoWiC}

\author{Ze Chen,\, Kangxu Wang,\, Zijian Cai,\, Jiewen Zheng,
        \bf Jiarong He,\, \bf Max Gao \\ \bf Jason Zhang \\
         Interactive Entertainment Group of Netease Inc., Guangzhou, China \\
         \texttt{\{jackchen,wangkangxu,caizijian01,zhengjiewen,gzhejiarong,jgao,}\\ \texttt{
         fyzhang\}@corp.netease.com}}

\begin{document}
\maketitle
\begin{abstract}
   This paper mainly describes the \textit{dma} submission to the TempoWiC task, which achieves a macro-F1 score of 77.05\% and attains the first place in this task. We first explore the impact of different pre-trained language models. Then we adopt data cleaning, data augmentation, and adversarial training strategies to enhance the model generalization and robustness. For further improvement, we integrate POS information and word semantic representation using a Mixture-of-Experts (MoE) approach. The experimental results show that MoE can overcome the feature overuse issue and combine the context, POS, and word semantic features well. Additionally, we use a model ensemble method for the final prediction, which has been proven effective by many research works.
\end{abstract}

\section{Introduction}
Lexical Semantic Change (LSC) Detection has drawn increasing attention in the past years\citep{liu2021-statistically,laicher2021explaining}. Existed research works \citep{liu2021-statistically} have shown that contextual word embeddings such as those produced by BERT \citep{devlin2018bert} have great advantages over non-contextual embeddings for inferring semantic shift when there is limited data. Meanwhile, many datasets are released to accelerate research in this direction. \citet{pilehvar2018wic} proposed Word-in-Context (WiC) dataset as an benchmark for generic evaluation of context-sensitive representations. \citet{raganato2020-xl} extended WiC to XL-WiC dataset with multilingual extensions.  
In contrast to these, TempoWiC \citep{TempoWiC} is crucially designed around the time-sensitive meaning shift and instances of word usage tied to Twitter trending topics. Our main work is to build a system that can detect semantic changes of target words in tweet pairs during different time periods for TempoWiC. It is framed as a binary classification task that addresses whether two instances of a target word have the same meaning. And pre-trained language models are adopted to produce contextual embeddings.

\section{Background}
\subsection{Task Description}
TempoWiC \citep{TempoWiC} is a new benchmark especially aimed at detecting a meaning shift in social media. Given a pair of sentences and a target word, the task is framed as a simple binary classification problem in deciding whether the meaning corresponding to the first target word in context is the same as the second one or not. 

The dataset of TempoWiC consists of 3297 annotated instances, which are divided into train/dev/test sets of size 1,428/396/1,473 instances, respectively. The target words involved in this task do not overlap between sets. For each sample, tweet pairs containing the target word were collected from the Twitter API at different time periods. The prior date is exactly one year before the peak date to avoid seasonal confound factors. The label True indicates that the word has the same meaning in the two tweets, while the label False indicates that the meaning is different.

\subsection{Pre-trained Language Models}
Recently, pre-trained language models (LM) have achieved remarkable achievement on natural language processing tasks, becoming one of the most effective methods for engineers and scholars. Transformers-based Pre-trained language models such as BERT\cite{devlin2018bert},  RoBERTa\cite{liu2019roberta}, DeBERTa\citep{he2020deberta}, DeBERTaV3\cite{he2021debertav3} is designed to pre-trained deep representation from unlabeled text, which can be fine-tuned with just one additional output layer to create state-of-the-art models for a wide range of tasks, such as question answering and language inference, without substantial task-specific architecture modifications.

By training language models using Twitter corpora from different time periods, \cite{loureiro2022timelms} showed that language undergoes semantic transformations over time, proving that training a language model with outdated corpora leads to a decline in performance.
\subsection{Mixture-of-Experts}
MoE \cite{arnaud2019treegated} is an approach for conditionally computing a representation. Given several expert inputs, the output of MoE is a weighted combination of the experts. Recently, MoE achieves significant improvements on several natural language processing tasks, such as named entity recognition \cite{meng2021gemnet}, recommendation\cite{zhu2020recommendation} and machine translation\citep{shazeer2017outrageously}.

\section{System Overview}
In this section, we first present the framework details for the models adopted in our work. Then we introduce several strategies for improving the models' robustness. Finally, we talk about the design of the model ensemble method.
\subsection{Models}
Our model framework can be divided into three layers:\textit{encoding}, \textit{matching} and \textit{prediction}. The \textit{encoding} layer is meant for sequence modeling to capture contextual semantic representation. The \textit{matching} layer focuses on finding out the interrelation and differences between the target words in two different tweets. And the \textit{prediction} layer is implemented as a classifier that decides whether the meaning of the target word is the same or not. 

\begin{figure}[htbp]
   \centering
   \includegraphics[width=\linewidth,scale=1.00]{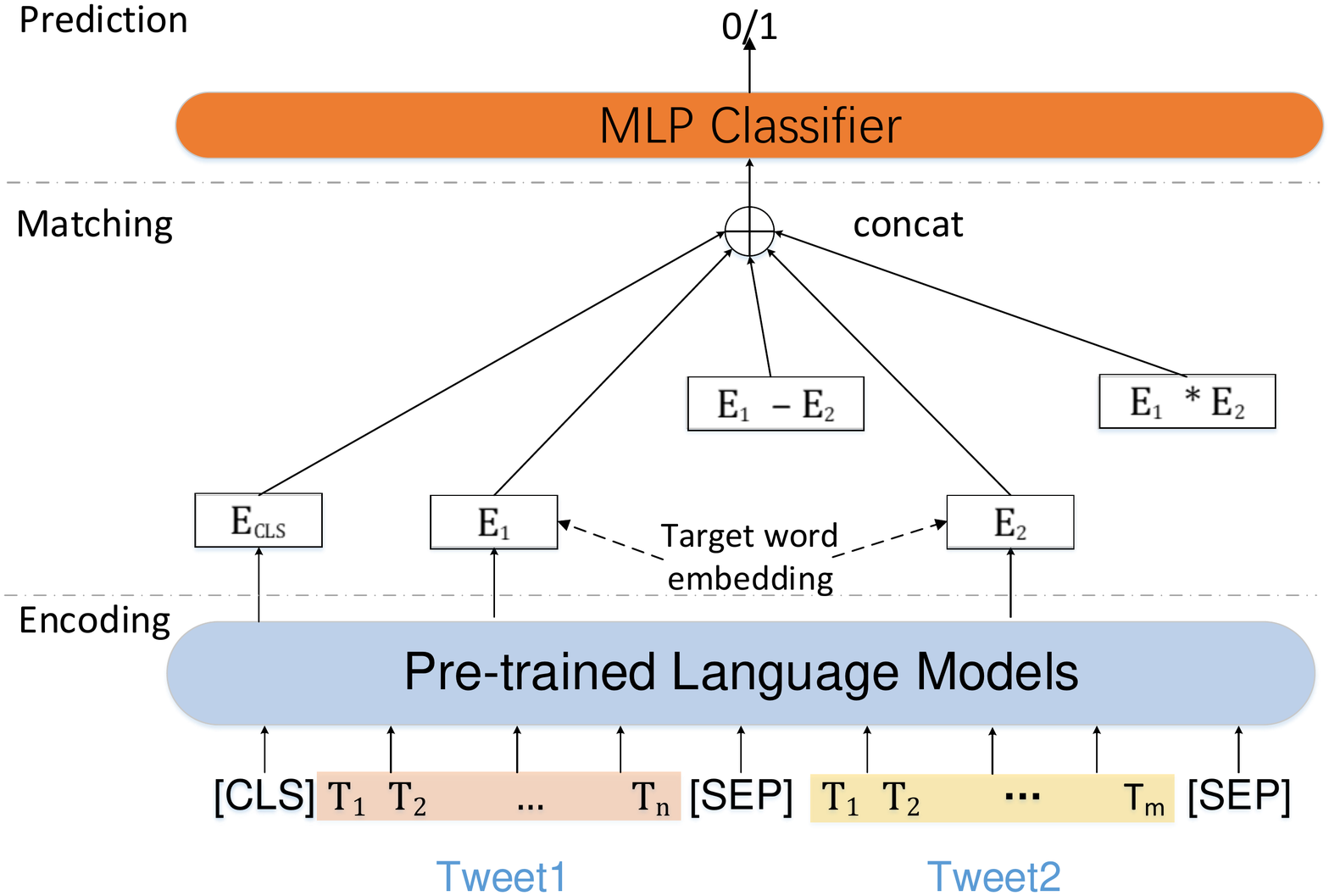} 
   \caption{Base Model Architecture}
   \label{base_model}
\end{figure}

\noindent\textbf{A. Base Model} \quad Figure \ref{base_model} shows the details of our base model. Two tweets are concatenated together and fed into a pre-trained LM, and the contextual embeddings(e.g. $E_1$, $E_2$ \footnote{We experimented with different target word representations: the first token in the word span, the mean value of all tokens in the span, the concatenation of the first token and last token in the span. And we found that adopting the concatenation of the first token and last token in the span can perform better than others. Please refer to Appendix \ref{appendA} for more details.}) corresponding to the target word on each tweet of the pair can be achieved. Then $E_1$ and $E_2$ are processed by the \textit{matching} layer to find the difference in these two tweets. The procedure can be summarized as follows:
$$E_{match}=[E_1; E_2; E_1-E_2; E_1*E_2; E_{CLS}]$$
$$y_{o} = softmax(MLP(E_{match}))$$
$$loss=CrossEntropy(y_{o}, y_{true})$$
where $E_{CLS}$ is the embedding of the first token, $E_{match}$ is the output of the \textit{matching} layer, $MLP$ is a multi-layer perceptron, $y_{true}$ is the gold label and $y_{o}$ is the output by the base model. $E_1*E_2$ means the Hadamard product of these two vectors, and $E_1-E_2$ represents the elementwise subtraction.

\begin{figure}[htbp]
   \centering
   \includegraphics[width=\linewidth,scale=1.00]{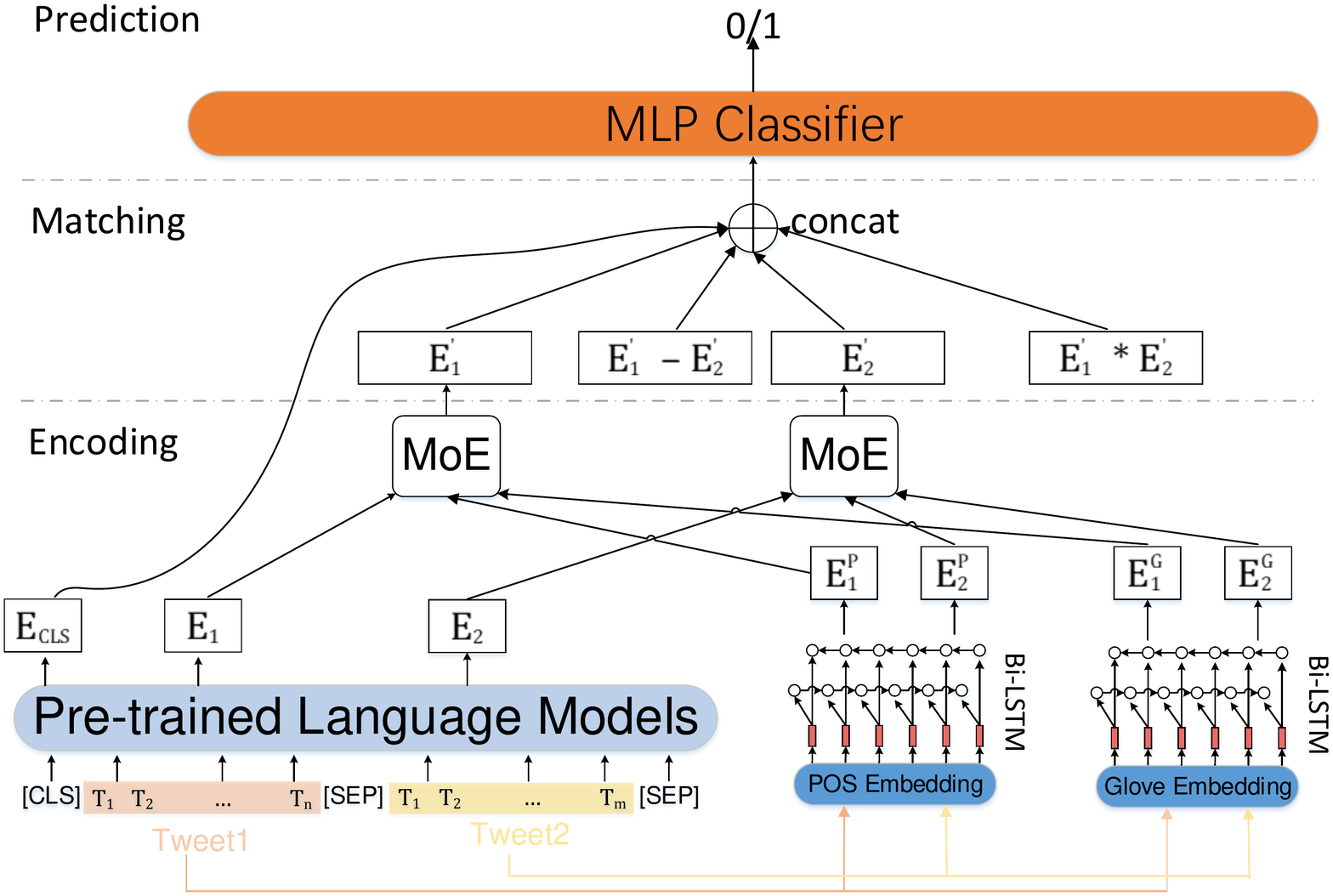} 
   \caption{MoE-based Model Architecture}
   \label{moe_model}
\end{figure}

\noindent\textbf{B. MoE Models} \quad Figure \ref{moe_model} gives a glimpse of our MoE-based model architecture. We extend the base model with two separate BiLSTM to integrate the POS information and the word semantic representation. For a pair of tweets, we first extract the contextual embeddings for the target word from pre-trained LM, and then we use two separate BiLSTM to get POS encoding and word semantic encoding. At last, an MoE module is adopted to merge these three encodings for the target word. The generated embeddings(e.g. $E^{'}_1$, $E^{'}_2$) for the target word are then processed by the \textit{matching} layer and \textit{prediction} layer as described above. Here we denote the POS encoding for the target word in the pair of tweets as $E^{P}_1,E^{P}_2$, and denote the GloVe-initialized word semantic encoding as $E^{G}_1, E^{G}_2$ respectively.

\begin{figure}[htbp]
   \centering
   \includegraphics[width=\linewidth,scale=0.8]{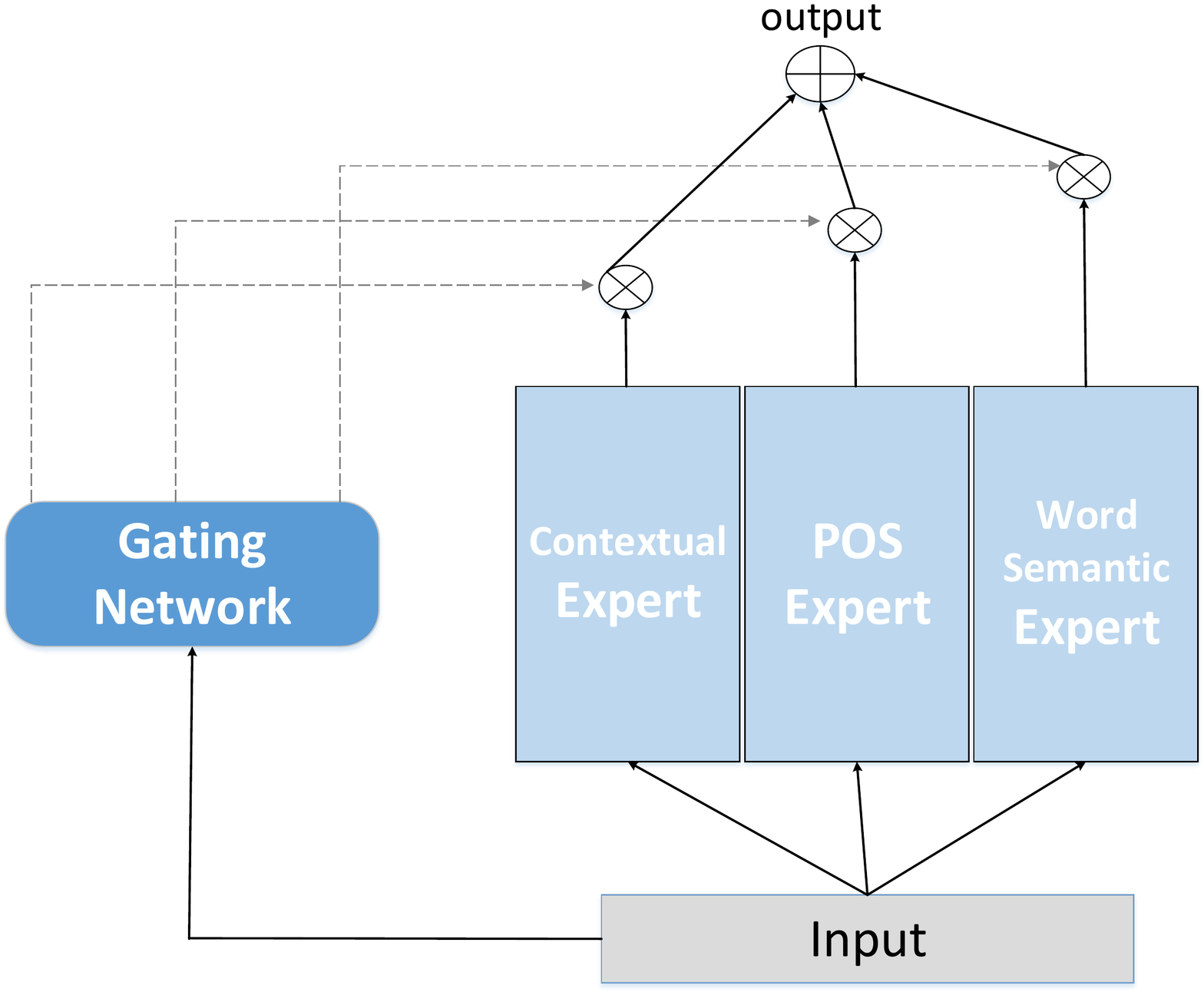} 
   \caption{Illustration of an MoE module}
   \label{moe}
\end{figure}

The details of an MoE module for this task are given in Figure \ref{moe}, which consists of a gating network and three experts. The procedure can be summarized as follows:
$$w_C, w_P, w_G = Gate(E_1, E^{P}_1, E^{G}_1)$$
$$E^{'}_1 = w_C*E_1 + w_P*E^{P}_1 + w_G*E^{G}_1$$
where $w_C, w_P, w_G$ are the weights for contextual expert, POS expert, and word semantic expert respectively, $Gate$ stands for the gating network, and $E^{'}_1$ is the output of MoE module for the target word in the first tweet. We can get $E^{'}_2$ for the target word in the second tweet by the same approach. And two gating networks are implemented here.

\begin{itemize}[leftmargin=*]
	\setlength{\itemsep}{0pt}
	\setlength{\parsep}{0pt}
	\setlength{\parskip}{0pt}
	\item {{\bf Separate Gating Network(S-Gate):} The weight for each expert is calculated separately. We define a task-specific vector $V_t$, the weight for expert $i$ can be calculated as: $w_i=\sigma(\theta[V_t, E^i])$, where $\theta$ are trainable parameters, $[,]$ is the concatenation and $\sigma$ is the Sigmoid activation, $E^i$ is the encoding of i-th  expert.}
	\item {{\bf Joint Gating Network(J-Gate):} The weights for all experts are calculated together. We define the weight vector for all experts as $W$, which is a three-dimension vector and can be calculated as: $W=softmax(\theta[E^1, E^2, E^3])$, where $\theta$ are trainable parameters.}
\end{itemize}

\subsection{Data Cleaning and Augmentation}
Given that the dataset is somewhat small, and there are some flaws in the labeled data, we adopt simple cleaning and augmentation strategies. We simply remove HTML tags and emojis in tweets, and replace the symbol \textit{@username} with a generic placeholder. Moreover, we directly remove the wrongly labeled samples of the target word position. There are many different data augmentation strategies: token shuffling, cutoff, back-translation, and so on. We just introduce the WiC dataset\citep{pilehvar2018wic} for data augmentation in this paper.

\subsection{Adversarial Training}
Adversarial attack has been well applied in both computer vision and natural language processing to improve the model's robustness. We implement this strategy with Fast Gradient Method\citep{goodfellow2014explaining}, which directly uses the gradient to compute the perturbation and augments the input with this perturbation to maximizes the adversarial loss. The training procedure can be summarized as follows:
$$\min_{\theta}E_{(x,y)\sim \mathcal{D}}[\max_{\Delta x\in \Omega} L(x+ \Delta x, y; \theta)]$$
where x is input, y is the gold label, $\mathcal{D}$ is the dataset, $\theta$ is the model parameters, $L(x+ \Delta x, y; \theta)$ is the loss function and $\Delta x$ is the perturbation.

\subsection{Model Ensemble}
For the final prediction, we implement a model ensemble method. In detail, we use one base model and the other two MoE models mentioned above to get the prediction scores and then average these output scores as the final result.

\section{Experiments}
\subsection{Experimental Setup}
Our implementation is based on the Transformers library by HuggingFace\citep{wolf2019huggingface} for the pre-trained models and corresponding tokenizers. During training, the data is processed by batches of size 8, the maximum length of each sample is set to 256, and the learning rate is set to 1e-6 with a warmup ratio over 10\%. By default, we set $\epsilon$ to 1.0 in FGM and set the MLP to two layers with a hidden size of 256.

When MoE models are employed, the hidden size of BiLSTM is set to 1024, and the pre-trained Twitter GloVe word vectors \footnote{\url{https://nlp.stanford.edu/projects/glove/}} are used for word embedding initialization. Moreover, we use nltk toolkit \footnote{\url{https://www.nltk.org/}} to extract POS tags, and the POS embeddings are randomly initialized. Our system jointly optimizes over different experts, but their model architectures differ. We adopt differential learning rates to tackle this problem. The learning rate for the transformer-based model is set to 1e-6, and the learning rate for BiLSTM is set to 1e-4.

\subsection{Results and Analysis}
In this section, we first present experimental results on the base model. Then we experiment with MoE models using the effective strategies validated on the base model. At last, the results of the model ensemble are reported.

We explore the impact of different pre-trained LMs adopted as the contextual encoder. Results given in Table \ref{baseResults} show that DeBERTa-large can perform well on this task. And TimeLMs\citep{loureiro2022timelms} can perform better than generic RoBERTa since they are implemented and adapted to the Twitter domain. Moreover, TimeLMs-2020-09 can achieve almost the best results among TimeLMs, largely because the dev dataset is distributed over this time period. From the last two rows in Table \ref{baseResults}, we can find that data cleaning and augmentation can increase the macro-F1 score by 2.83 percentage points, and FGM training can increase this indicator by 2.57\%. Additionally, the ablation study results on \textit{matching} layer are presented in Appendix \ref{appendA}, we can find that the first token $[CLS]$ embedding can help improve the performance of this task. The subtraction and Hadamard product operations can also help find the difference between target words in two tweets. 

When we experiment with MoE models, the data cleaning and augmentation, and FGM training are adopted by default. And the pre-trained DeBERTa-large is used for the contextual encoder. Table \ref{MoEResults} shows the performance of different MoE models. We can find that when integrating POS information and word semantic representation by using an MoE architecture, the performance can improve a lot. More specifically, the MoE model with S-Gate and J-Gate can achieve macro-F1 scores of 79.25\% and 79.19\% respectively, both of which increase the base by more than 2\%. For further analysis, ablation studies are done here. We experiment with POS information and GloVe separately and find that using an MoE model to integrate POS information can improve the performance by 1\%, while using an MoE model to combine word semantic representation can increase the macro-F1 score by about 2\%.

Table \ref{EnsembleResults} gives the results of our model ensemble method. By averaging the prediction scores of one base model and the other two MoE models(S-Gate + POS + GloVe, J-Gate + POS + GloVe), the macro-F1 score can increase by more than 1\% on the dev dataset. And our model ensemble method achieves a macro-F1 score of 77.05\% on the test dataset, which attains the first place in this task.

\begin{table}\footnotesize
\centering
\begin{tabular}{ccc} 
\hline
\textbf{Model}  & \textbf{Accuracy} & \textbf{macro-F1} \\ 
\hline
TimeLMs-2019-12  & 67.17\%            & 63.34\% \\
TimeLMs-2020-03  & 68.18\%            & 65.20\% \\
TimeLMs-2020-09  & 68.43\%            & 65.42\% \\
TimeLMs-2020-12  & 68.18\%            & 65.20\% \\
TimeLMs-2021-03  & 66.67\%            & 63.88\% \\
TimeLMs-2022-03  & 68.18\%            & 65.12\% \\
\hline
\hline
RoBERTa-base & 61.62\%            & 60.30\% \\
DeBERTa-base & 69.90\%            & 65.60\% \\
DeBERTa-large & 71.72\%            & 71.68\% \\
+ Data Aug & 74.63\%            & 74.51\% \\
+ Data Aug + FGM(Base) & \bf{77.53}\%            & \bf{77.08}\% \\
\hline
\end{tabular}
\caption{Results of base model on dev dataset}
\label{baseResults}
\end{table}

\begin{table}\footnotesize
\centering
\begin{tabular}{lcc} 
\hline
\textbf{Model}  & \textbf{Accuracy} & \textbf{macro-F1} \\ 
\hline
Base  & 77.53\%          & 77.08\% \\
\hline
\hline
S-Gate + POS + GloVe          & \bf{79.29}\%      & \bf{79.25}\% \\
S-Gate + POS         & 78.26\%      & 77.20\% \\
S-Gate + GloVe         & 78.19\%      & 77.18\% \\
J-Gate + POS + GloVe    & \bf{79.29\%}   & 79.19\%   \\
J-Gate + POS         & 78.62\%      & 78.31\% \\
J-Gate + GloVe         & 77.55\%      & 77.12\% \\
\hline
\end{tabular}
\caption{Results of MoE-based models on dev dataset}
\label{MoEResults}
\end{table}

\begin{table}[H]
\centering
\begin{tabular}{ccc} 
\hline
\textbf{Dataset}  & \textbf{Accuracy} & \textbf{macro-F1} \\ 
\hline
Dev          & 80.81\%      & 80.5\% \\
Test    & \bf{78.34\%}   & \bf{77.05\%}   \\
\hline
\end{tabular}
\caption{Ensemble results on both Dev and Test dataset}
\label{EnsembleResults}
\end{table}

\section{Conclusion}
In this work, we provide an overview of the combined approach to detect the meaning shift in social media. We investigate the impact of adopting different pre-trained LMs, finding that DeBERTa performs best for this task. Experimental results show that strategies such as data augmentation and adversarial training can enhance the model's robustness. In particular, incorporating POS information and word-level semantic representation with MoE models can significantly improve performance. For future work, we will investigate how to incorporate different TimeLMs with MoE models for this task.

\bibliography{anthology,custom}
\bibliographystyle{acl_natbib}

\appendix

\section{Additional Experiments on the base model}
\label{appendA}
In this part, we present several additional experimental results on the base model.

We tried different target word representation methods for contextual embedding. The results on dev dataset are listed in Table \ref{Apd1}.
\begin{table}[H]\footnotesize
\centering
\begin{tabular}{lcc} 
\hline
\textbf{Target word}  & \textbf{Accuracy} & \textbf{macro-F1} \\ 
\hline
$First$          & 75.6\%      & 74.49\% \\
$Mean$    & 74.94\%   & 74.46\%   \\
$First$ + $Last$    & \bf{77.53\%}   & \bf{77.08\%}   \\
\hline
\end{tabular}
\caption{Results of different target word representation methods}
\label{Apd1}
\end{table}

To make further analysis, we conducted ablation studies to investigate the contribution of different components of \textit{matching} layer. Results are shown in Table \ref{Apd2}.
\begin{table}[H]\footnotesize
\centering
\begin{tabular}{rcc} 
\hline
\textbf{\textit{Matching} layer}  & \textbf{Accuracy} & \textbf{macro-F1} \\ 
\hline
$E_1$ + $E_2$  & 75.92\%          & 74.74\% \\
+ $E_{CLS}$  & 77.15\%          & 76.45\% \\
+ $E_{CLS}$ +[$E_1$-$E_2$] + [$E_1$*$E_2$]  & \bf{77.53}\%          & \bf{77.08}\% \\
\hline
\end{tabular}
\caption{Results of different components of \textit{matching} layer}
\label{Apd2}
\end{table}

\end{document}